\documentclass[11pt]{article}

\usepackage{amsmath,amsthm,amsfonts,mathrsfs,anysize,fancyhdr,epsfig,mdframed, float,graphicx,umoline,dsfont}
\newtheorem{theorem}{Theorem}

\usepackage{mathtools}                                                      
\mathtoolsset{showonlyrefs=true}                                    
\usepackage{fixltx2e,amsmath}                                           
\MakeRobust{\eqref}

\providecommand{\keywords}[1]{\textbf{\textit{Keywords}} #1}                                        
\begin{document}

\title{A Gaussian Process perspective on Convolutional Neural Networks}
\author{Anastasia Borovykh\thanks{Dipartimento di Matematica, Universit\`a di Bologna, Bologna, Italy.\textbf{e-mail}: borovykh\_a@hotmail.com}
}
\date{This version: \today}
 
 \maketitle 
 \begin{abstract}
In this paper we cast the well-known convolutional neural network in a Gaussian process perspective. In this way we hope to gain additional insights into the performance of convolutional networks, in particular understand under what circumstances they tend to perform well and what assumptions are implicitly made in the network. While for fully-connected networks the properties of convergence to Gaussian processes have been studied extensively, little is known about situations in which the output from a convolutional network approaches a multivariate normal distribution.
 \end{abstract}
 

\section{Introduction}
A convolutional neural network (CNNs) is a biologically-inspired type of deep neural network (DNN) that has recently gained popularity due to its success in classification problems in particular computer vision and speech recognition, see e.g. \cite{krizhevsky12}, \cite{kim14}, \cite{karpathy14}, but also in forecasting problems \cite{vanoord16}. The CNN consists of a sequence of convolutional layers, the output of which is connected only to local regions in the input. This is achieved by a convolution: sliding a filter over the input and at each point computing the dot product between the two. This structure allows the model to learn filters that are able to recognize specific patterns in the input data and moreover combine the learned patterns in a hierarchical way. 

While convolutional networks have been shown to possess the ability of learning invariant features, many of the inner workings of a CNN still remain unknown. Approaching the convolutional network from a Bayesian perspective by assuming a probability distribution on the network parameters one can induce a distribution on the output function. While computing the distribution of the output of the network exactly can be challenging, one can study the network by means of its limiting behavior.

It is well known that a neural network with an infinite number of hidden nodes and weights initialized from a Gaussian distribution, by an application of the central limit theorem (CLT), outputs a function with a Gaussian process prior \cite{neal12}. This work motivated a theoretical understanding of neural networks and deep learning through the kernel methods and in particular Gaussian processes (GPs).

\subsection{Our contributions}
While convolutional neural networks were developed from a practice-based approach, and have empirically been shown to be very powerful, in this work we extend the theoretical understanding of CNNs by connecting convolutional neural networks to Gaussian processes. Unlike fully-connected neural networks (FNN) in which the output layer computes a sum of independent and identically distributed random variables, in the convolutional network the sum in each layer is computed over uncorrelated and not necessarily identically distributed weighted variables from the previous layer, thus possessing a compositional structure which the fully-connected net lacks. While the general CLT is unapplicable in this non-identically distributed case, we can employ a generalized central limit theorem \cite{bentkus05} in the form of a Lyapunov-type bound to study the convolutional network from a Gaussian process perspective, at least in the first layer. Our work therefore extends the well-known concept of convergence to GP behavior in wide fully-connected neural networks to convolutional networks. We then study the discrepancy between the two numerically and show that already for relatively small filter widths, even if the CLT does not fully apply for deeper networks, the output approaches a Gaussian process relatively quickly. 

\subsection{Related work}
Gaussian processes have been a well-known instrument used in forecasting time series by defining through them a flexible prior over functions in regression models \cite{rasmussen06}. The ability of the GP to learn meaningful dependencies is fully encoded by the covariance function, a function that specifies the dependence between the inputs and  learning a flexible and expressive covariance kernel is therefore an active topic of research, see e.g. \cite{gibbs98}, \cite{saul09}, \cite{wilson13}. The Gaussian process uses infinitely many fixed basis functions, and typically work as smoothing devices, as opposed to the neural network which use a finite number of adaptive basis functions. Combining the non-parametric flexibility of the GPs with the adaptiveness of neural networks can therefore improve the generalisation capabilities of the GPs, see e.g. \cite{hinton08}, \cite{wilson11}, \cite{bradshaw17}. In particular, the authors in \cite{wilson16} used a deep neural network to transform inputs into complex kernels functions, in which the DNN parameters become kernel hyperparameters and \cite{calandra16} used a neural network to transform the input space into a feature space and training a Gaussian process on this space. In \cite{rasmussen17} the authors use a convolutional structure in combination with Gaussian processes to create a convolutional kernel and improve the generalisation abilities; the difference with our work here being that we study the ability of the CNN to by construction perform as a GP. The convolutional structure also has similarities with the additive Gaussian processes \cite{duvenaud11}, \cite{durrande12}. These deep and additive kernels bear similarities with the kernel of the Gaussian process that naturally arises when relating the GP to the CNN. As we will show, it is defined as a sum of random variables, in which the kernel in the final output layer can be recursively defined through the previous layer kernel functions.

The relation between neural networks with one hidden layer and GPs was studied in \cite{neal12}. In particular cases of the activation function the resulting kernels can be computed analytically, see \cite{williams97} where analytic expressions for the sigmoidal and Gaussian activations are derived. In \cite{saul09} the authors provided analytical expressions for the recursive kernel dependency to mimic the behavior in a deep network when using a rectified linear unit activation function. Similarly, the work of \cite{hazan15} studied the behaviour of the GP kernels in deep infinite networks. In \cite{lee17} the authors identify the relation between using these kernels as the covariance function for a GP and predicting with Bayesian deep neural networks. The authors of \cite{matthews18} also study the connection between deep fully connected networks to GPs, in particular focussing on the convergence of the deep finite networks to Gaussian processes and the discrepancy between their outputs. Also worth mentioning is the area of deep Gaussian processes \cite{damianou13} which essentially stack several GPs giving rise to a richer class of probability models.

\section{Background}
Gaussian processes gained popularity in the machine learning community after the work of \cite{neal12} showed that the neural network prior tends to a Gaussian process as the number of hidden units tends to infinity. In this section we shortly repeat the main conclusions. 
\subsection{Gaussian processes}\label{sec21}
Consider the $n$-dimensional vector of function values $(f(x^1),...,f(x^n))$ evaluated at $x^i\in\mathcal{X}$ where $\mathcal{X}$ is the input space, $i=1,...,n$. Then for $f:\mathcal{X} \rightarrow \mathbb{R}$ we say that $f$ is a Gaussian process \cite{rasmussen06}, i.e.
\begin{align}
f(x) \sim \mathcal{GP}(m(x), k(x,x')),
\end{align}
if any finite subset $(f(x^1),...,f(x^n))$ has a multivariate Gaussian distribution. The Gaussian process is defined over the index set $\mathcal{X}$ here equivalent to the input domain, and it is completely specified by its mean function and covariance function as
\begin{align}
m(x) &= \mathbb{E}(f(x)),\\
k(x,x') &=\textnormal{cov}(f(x),f(x'))\\
&= \mathbb{E}((f(x) - m(x))f(x')-m(x'))),
\end{align}
for $x,x'\in\mathcal{X}$. The covariance function $k:\mathcal{X}\times\mathcal{X}\rightarrow\mathbb{R}$ defines the nearness or similarity between two inputs $x$ and $x'$. 
Given a sample of input points $X=(x^1,...,x^n)$, the covariance matrix for this sample $K(X,X)\in\mathbb{R}^{n\times n}$ has entries $K_{i,j}=k(x^i,x^j)$. Consider now the set of training points $\mathcal{D} = \{(x^i,y^i)_{i=1}^N\}$. Let $y = f(x) + \epsilon$, with a Gaussian likelihood, i.e. $\epsilon\sim\mathcal{N}(0,\sigma_\epsilon^2)$ and assume a Gaussian prior on the function $f\sim \mathcal{GP}(0,k)$. If the prior on $f$ is a GP and the likelihood is Gaussian the posterior on $f$ is also a GP and the predictive distribution is given by
\begin{align}\label{eq:posterior}
p(y_*|x_*,\mathcal{D}) &= \int p(y_*|x_*,f,\mathcal{D})p(f|\mathcal{D})df\\
&=\mathcal{N}(\bar f_*,\textnormal{cov}(f_*)),\\
\bar f_* &= K(x_*,x)(K(x,x)+\sigma_\epsilon^2 I)^{-1}y,\\
\textnormal{cov}(f_*) &= K(x_*,x_*)-K(x_*,x)(K(x,x)+\sigma_\epsilon^2I)^{-1}K(x,x_*).
\end{align}


\subsection{Neural networks and Gaussian processes}
Consider an input $x\in\mathcal{X}$ and a fully-connected neural network consisting of $L\geq 2$ layers with $M$ hidden nodes in each layer $l=1,...,L$. 
Each layer $l$ in the network then computes for each $i=1,...,M$ 
\begin{align}
a_i^l(x) = \sum_{j=1}^{M} w^l_{i,j}z^{l-1}_j + b^l_j,\;\;\;z_i^l(x) = h(a_i^l(x))),
\end{align}
where we explicitly denote the dependence on a particular input sample $x$, and $w^l\in\mathbb{R}^{M\times M}$, $b^l\in\mathbb{R}^M$ and $h(\cdot)$ is the nonlinear activation function. Note that in the first layer we compute the linear combination using the input, i.e. $z^0 = x$ and $M=d$. 
From \cite{neal12}, we know that a neural network with the number of hidden units tending to infinity approaches a Gaussian process prior. Let the weights $w^l$ and biases $b^l$ have zero mean Gaussian distributions with variances $\sigma_w^2=\nu_w^2/M$ and $\sigma_b^2$, respectively. If the units $a_i^{l-1}$ are independent for different $i$, then $a_i^l$ is a sum of i.i.d. terms -- the weights and hidden units -- and from the CLT it follows that in the limit of $M\rightarrow \infty$, the $a_i^l$ will be normally distributed. Similarly consider the joint distribution of $(a_i^l(x^{1}), ..., a_i^l(x^{N}))$, where $(x^{1},...,x^{N})$ is a finite set of input samples. Since each unit consists of a sum i.i.d. terms with zero mean and fixed covariance, we can apply the multivariate CLT. This tells us that this distribution is multivariate Gaussian, so that
\begin{align}
a_i^l &\sim \mathcal{GP}(0,k^l(x^{p},x^{q})),\\
k^l(x^{p},x^{q}) &:=\mathbb{E}\left[a_i^l(x^{p})a_i^l(x^{q})\right]\\
&=\sigma_b^2+\sum_{j=1}^M\sigma_w^2\mathbb{E}\left[z^{l-1}_j(x^{p})z_j^{l-1}(x^{q})\right]\\
&= \sigma_b^2+\nu_w^2C^l(x^{p},x^{q}),
\end{align}
 with $C^l(x^{p},x^{q})=\mathbb{E}\left[h(a^{l-1}_j(x^{p}))h(a_j^{l-1}(x^{q}))\right]$ the same for all $j$. Ways of analytically computing this kernel are discussed in \cite{williams97}, while the authors in \cite{lee17} present a numerical way of computing the kernel for deep neural networks. We note here that the units $a_i^{l}$ for different $i$ are uncorrelated, but since they are only multivariate normal in the limit, they are not necessarily independent for finite layer widths. This means that as we add more layers to the network, the output will deviate from a Gaussian due to more dependency being introduced in the terms of the summation; this is similar to what is mentioned in the work of \cite{matthews18}.

\section{Convolutional neural networks and Gaussian processes}
Consider a convolutional neural network in a one-dimensional setting for ease of notation. Suppose we are given a sequence of inputs $X=(x_1,...,x_d)$, where each $x_i\in\mathbb{R}$ is assumed to be one-dimensional and a set of outputs $Y=(y_1,...,y_{d'})$.  The output from the first layer is given by convolving the filter $w^1$ with finite support with the input:
\begin{align}\label{eq:conv1}
a^1_i = (w^1 * x)(i) = \sum_{j=1}^M w^l_j x_{i-j},\;\; z^1 = h(a^1),
\end{align}
where $w^1 \in\mathbb{R}^{1\times M}$ and $a^1\in\mathbb{R}^{1\times N-M+1}$ and the linear combination is passed through the non-linearity $h(\cdot)$ to give $z^1 = h(a^1)$. In each subsequent layer $l=2,...,L$ the input feature map, $z^{l-1}\in \mathbb{R}^{1\times N_{l-1}}$, where $1\times N_{l-1}$ is the size of the output filter map from the previous convolution with $N_{l-1} = N_{l-2}-M+1$, is convolved with a filter $w^l\in\mathbb{R}^{1\times M}$, to create a feature map $a^l\in\mathbb{R}^{1\times N_l }$:
\begin{align}\label{eq:conv2}
a^l_i = (w^l * z^{l-1})(i) = \sum_{j=1}^M w^l_jz^{l-1}_{i-j},\;\; z^l_i=h(a^l_i),
\end{align}
with the forecasted output given by $\hat y=a^L$. Define the receptive field to be the number of inputs which modify the output. The filter size parameter $M$ thus controls the receptive field of each output node. Without zero padding, in every layer the convolution output has width $N_l = N_{l-1}-M+1$ for $l=1,..,L$ and by padding the input with a vector of zeros the size of the receptive field we can control the output size to have the same size as the input. Since all the elements in the feature map share the same weights this allows for features to be detected in a time-invariant manner, while at the same time it reduces the number of trainable parameters. 
\begin{figure}[h]
\centering
\includegraphics[scale=0.3]{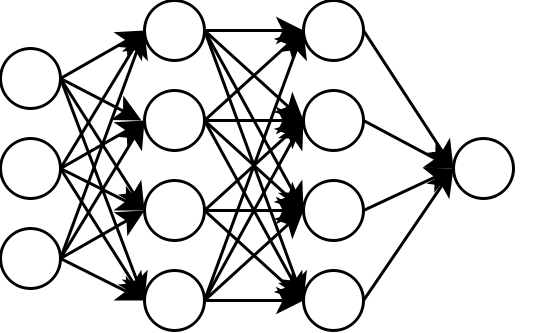}\vspace{0.5cm}
\includegraphics[scale=0.25]{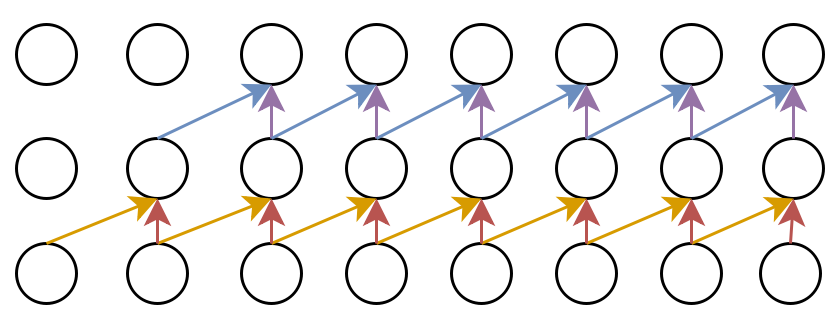}
\caption{A configuration of a two-layer fully-connected network (left) and a convolutional network with a filter size of two, the weights are shared across an input layer as indicated by the similar colors (right).}\label{fig:fig1}
\end{figure} 

\subsection{Central limit theorem}\label{sec31}
Consider now the convolutional neural network given in \eqref{eq:conv1} - \eqref{eq:conv2}. Let the inputs $(x_1,...,x_T)$ be the input vector. Initialize all weights in the convolutional neural network to be independent samples from a Gaussian prior $\mathcal{N}(0,\sigma_w^2)$. 
Consider the output of the convolution in the first layer $a^1\in\mathbb{R}^d$ where we let $d$ be the placeholder for the dimension of the output. We have $a^1 = [a_1^1,a_1^2,...,a^1_d]^T$ with covariance matrix
\begin{align}
K^1=\begin{bmatrix}
\mathbb{E}[a_1^1a_1^1]&\mathbb{E}[a_1^1a_2^1]&\cdots&\mathbb{E}[a_1^1a_d^1]\\
\mathbb{E}[a_2^1a_1^1]&\mathbb{E}[a_2^1a_2^1]&\cdots&\vdots\\
\vdots&\vdots&\ddots&\vdots\\
\mathbb{E}[a_d^1a_1^1]&\cdots&\cdots&\mathbb{E}[a_d^1a_d^1]
\end{bmatrix},
\end{align}
in which
\begin{align}
a^1_i=\sum_{j=1}^Mw_j^1x_{i-j}, \;\; K^1_{i,k}=\sigma_w^2\sum_{j=1}^M\mathbb{E}[x_{i-j}x_{k-j}]\label{eq:cov1}.
\end{align}
The matrix $K^1$ thus defines the covariance between the elements of the vector $a^1$. 
Conditional on the input $x$ the terms in the summation for $a^1$ are independent by the independence between the input and the weights and the i.i.d. distribution of the weights.
However, the terms are not necessarily identically distributed. In particular, we have $E[w_j^1x_{i-j}]=0$ for all $i$ but $\mathbb{E}[w_j^1x_{i-j}w_j^1x_{i-j}]=\sigma_w^2\mathbb{E}[x_{i-j}x_{i-j}]$ does not have to equal $\sigma_w^2\mathbb{E}[x_{i-k}x_{i-k}]$. In vector notation
 \begin{align}
 V_1^1=\begin{bmatrix}
w_1x_{1-1}\\w_1x_{2-1}\\...\\w_1x_{d-1}
\end{bmatrix},\;\;
...\;\;,\;\;
V_M^1=\begin{bmatrix}
w_Mx_{1-M}\\w_Mx_{2-M}\\...\\w_Mx_{d-M}
\end{bmatrix},
 \end{align}
 where each vector $V_j^1$ has covariance matrix with elements $\Sigma_{i,k} = \sigma_w^2 x_{i-j}x_{k-j}$. Conditionally on the inputs these vectors are independent but not identically distributed. Therefore, a straightforward application of the multivariate central limit theorem, as is done in the fully-connected neural network case, does not apply due to the elements in the summation not being identically distributed. 
 
 Let $|x|^2=x_1^2+\cdots+x_d^2$. We introduce the following Theorem as proven in \cite{bentkus05} 
\begin{theorem}[A Lyapunov-type bound in $\mathbb{R}^d$]\label{theorem1}
Let $X_1,\cdots,X_M$ be independent random vectors taking values in $\mathbb{R}^d$ such that $\mathbb{E}[X_i]=0$ for all $i$. Let $S=X_1+\cdots+X_M$. Assume that the covariance operator $\Sigma^2$ of $S$ is invertible. Let $Z\sim\mathcal{N}(0,\Sigma^2)$, a centered Gaussian with covariance matrix $\Sigma^2$. Let $\mathcal{C}$ stand for the class of all convex subsets of $\mathbb{R}^d$. Then
\begin{align}\label{eq:bound}
\sup_{A\in\mathcal{C}}\bigg|\mathbb{P}(S\in A)-&\mathbb{P}(Z\in A)\bigg|\\
&\leq \mathcal{O}(d^{1/4})\sum_{i=1}^M\mathbb{E}\left[\big|\Sigma^{-1}X_i\big|^3\right].
\end{align}
\end{theorem}
In particular the above Theorem is able to give a rate of convergence between the random variables $S$ and $Z$ but does not require the random variables in the sum to be identically distributed, making it well-suited for application in the convolutional neural network. As a special case of the above we have
\begin{theorem}[Independent and identically distributed case]\label{theorem2}
If the random variables $X_1,\cdots,X_M$ are i.i.d. and have identity covariance $\mathbb{I}_d$, then the lower bound specifies to 
\begin{align}
\sup_{A\in\mathcal{C}}\bigg|\mathbb{P}(S\in A)-&\mathbb{P}(Z\in A)\bigg|\\
&\leq \mathcal{O}(d^{1/4})\mathbb{E}\left[\big|X_1\big|^3\right]/\sqrt{M}.
\end{align}
\end{theorem}
Note that the assumption $\Sigma^2 = \mathbb{I}_d$ is not restrictive since we can rescale both $S$ and $Z$ by $\Sigma^{-1}$ to give the identity covariance matrix. In other words, the above Theorems give a bound on the distance between the probability distribution of the (vector-valued) sum of not necessarily identically distributed variables, i.e. the output $a^l$ in layer $l$ where $l=1,2,...$ as computed in a convolutional neural network, and the probability distribution of a multivariate Gaussian random variable, i.e. the output of a Gaussian process in which the kernel is given by $\Sigma^2$. In order to be able to measure the closeness of the CNN and the GP we thus need to evaluate the right-hand side of equation \eqref{eq:bound}. Note that the closeness between the two is thus determined by the kernel function $\Sigma^2$, which as we will see in the next section, depends on the output in the particular type of activation function used and has a recurrent dependency on the outputs in previous layers.

\subsection{Gaussian process convergence}
By Theorem \ref{theorem1}, as the filter size tends to infinity the output from the first layer, $a^1$, follows a multivariate normal distribution and has a covariance matrix with elements given in \eqref{eq:cov1}. For the output and covariance matrix in the next layers we have
\begin{align}
a_i^l = \sum_{j=1}^Mw_j^lz^{l-1}_{i-j}, \;\; K^l_{i,k} = \sigma_w^2\sum_{j=1}^M\mathbb{E}\left[z^{l-1}_{i-j}z^{l-1}_{k-j}\right]\label{eq:fingp}
\end{align}
Assume $a^{l-1}$ follow a multivariate normal distribution. The terms in the summation of $a^l_i$ are are uncorrelated due to the weights being sampled i.i.d.,
\begin{align}
\mathbb{E}[w_j^lz_{i-j}^{l-1}w_k^lz_{i-k}^{l-1}]=\mathbb{E}[w_j^1w_k^1]\mathbb{E}[z_{i-j}^{l-1}z_{i-k}^{l-1}]=0.
\end{align}
In the case of a multivariate normal distribution a set of variables having a covariance of zero implies that the variables are mutually independent. The problem here comes from the fact that a product of the weights with the $z^{l-1}$ is not a multivariate normal distribution and we cannot conclude independence as such. We can however hypothesize that similar to Theorem 1 in \cite{matthews18} there exist functions $h_1(M),...,h_l(M)$ such that the network output converges to a Gaussian process. We verify such a convergence numerically in Section \ref{sec4}.

\subsection{Kernel specification} The correlation between nodes in layer $l$ is a sum over the correlations in the receptive field of size $M$ in the previous layer. In this way the convolutional network, as opposed to the neural network, `averages' the covariances. Furthermore, due to the summation being dependent on the $M$ outputs in the previous layer, which in turn depend on $M$ outputs in the layer before that, and so forth, we see that the convolutional kernel structurally differs from the kernel obtained in the fully-connected neural network in the sense that the kernel in layer $l$ is dependent on a number of input terms that exponentially grows with $l$. In other words, the receptive field of a convolutional neural network determines the number of terms of the input on which the kernel is dependent.

The above covariance function bears similarities with the convolutional kernel used in\\ \cite{rasmussen17}, see e.g. equation (14) of \cite{rasmussen17} where the kernel function for two vector-valued inputs $x$ and $x'$ is given by
\begin{align}
K(x,x') = \sum_{i=1}^P\sum_{j=1}^P k(x_i,x'_j),
\end{align}
with $x_i$ indicating the $i$-th element of the vector $x$, $P$ is the number of patches or terms we sum over in the vector and $k$ is some underlying kernel, the choice of which would depend on the particular data structure. The difference with our result is that we show that such an additive kernel arises naturally in a convolutional neural network, if the CNN output prior is sufficiently close to that of the GP, and in our case the underlying kernel is determined by the activation function used. Furthermore, the number of patches we sum over is determined by $M$, the filter size, but also $l$, the number of layers in the network due to the recurrent dependence on the previous layer output in the kernel function. 
 
The rectified linear unit (ReLU) is commonly used as non-linearity in between the layers in both classification and forecasting tasks. Assuming that each layer output is close to a multivariate distribution, the expected value in \eqref{eq:fingp} is over the bivariate normal distribution of $z^{l-1}_{i-j}$ and $z^{l-1}_{k-j}$ with mean zero and covariance matrix with elements $K^l_{i-j,k-j}$, $K^l_{k-j,i-j}$, $K^l_{i-j,i-j}$ and $K^l_{k-j,k-j}$. For the ReLU non-linearity the 
expected value can be computed analytically, see \cite{saul09}. We thus obtain the recursive connection (similar to the recursive relationship derived in equation (12) and (13) for a fully-connected network in \cite{saul09})
\begin{align}
K^l_{i,k} = &\frac{\sigma_w^2}{2\pi}\sum_{j=1}^M\sqrt{K^{l-1}_{i-j,i-j}K^{l-1}_{k-j,k-j}}\\
&\cdot \left(\sin\theta_{i-j,k-j}^{l-1}+(\pi-\theta^{l-1}_{i-j,k-j})\cos\theta_{i-j,k-j}^{l-1}\right),\\
\theta^l_{i,k} =& \cos^{-1}\left(\frac{K^{l}_{i,k}}{\sqrt{K^l_{i,i}K^l_{k,k}}}\right),\label{eq:kernelgp}
\end{align} 
where we again remark on the summation over the receptive field being the difference with the neural network and the starting point $K^1$ is given in \eqref{eq:cov1}.

\begin{figure}[h]
   \centering
   \includegraphics[width=0.4\textwidth]{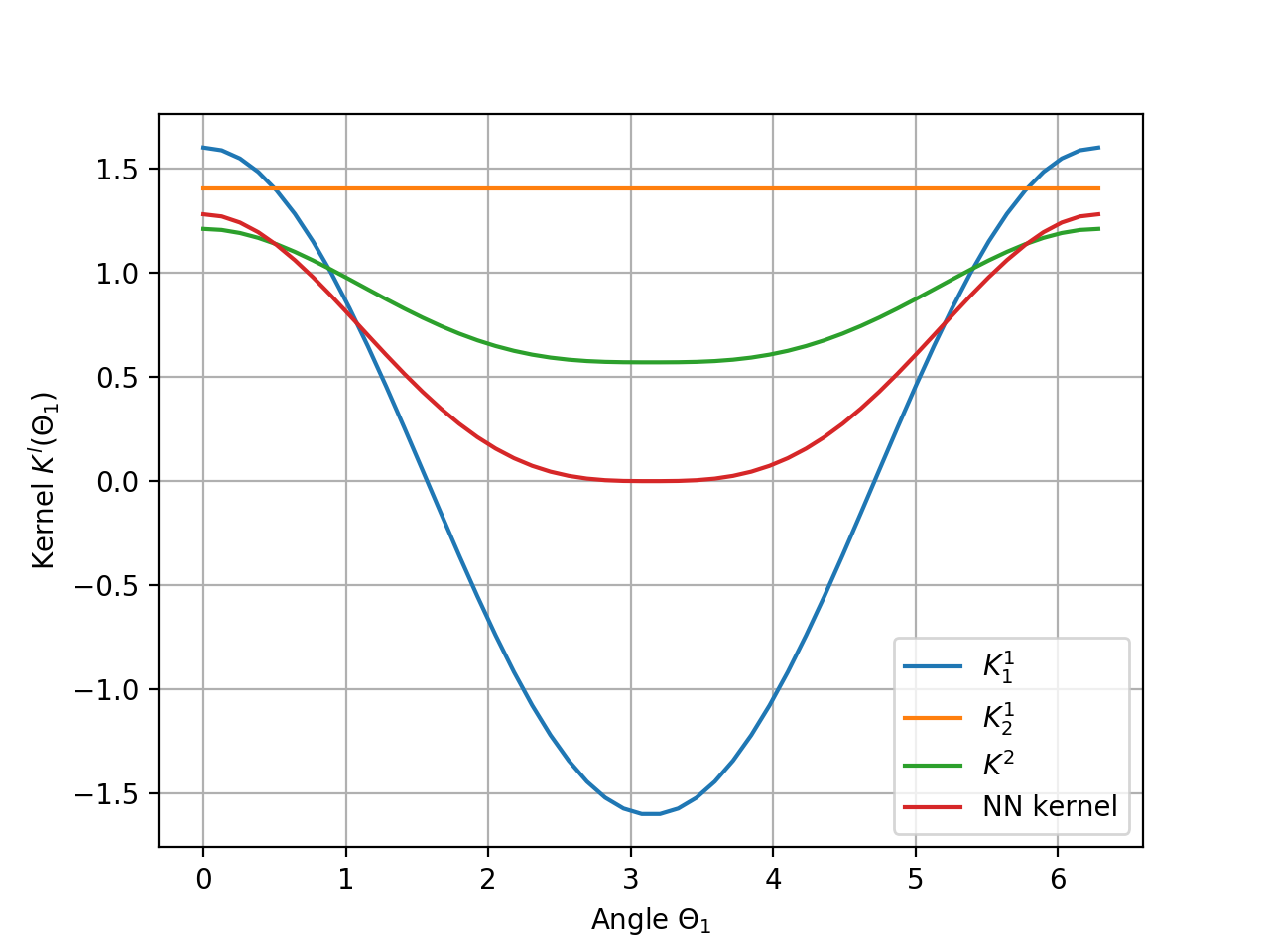}
      \includegraphics[width=0.4\textwidth]{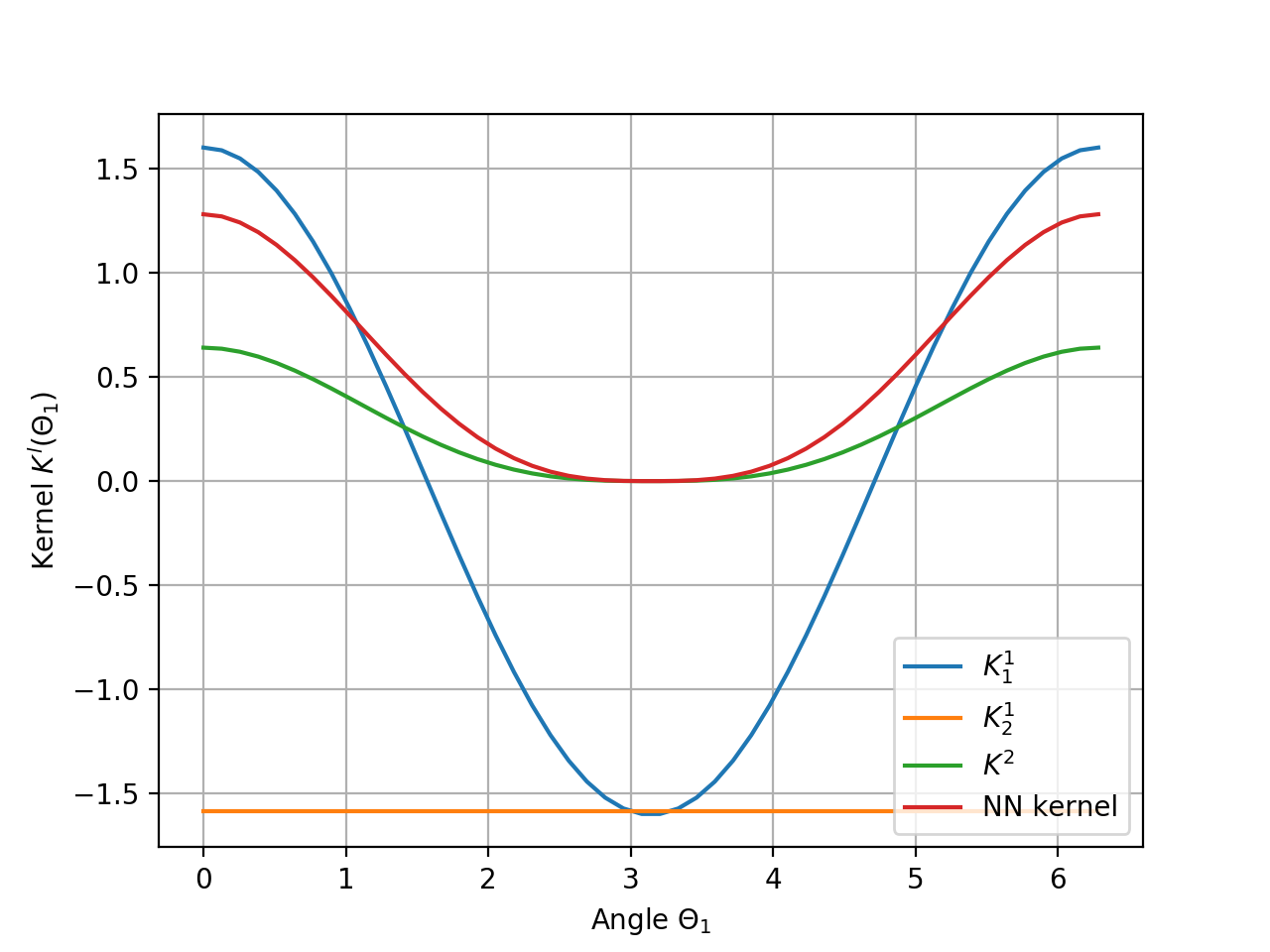}
   \caption{The angular structure of the convolutional kernel as a function of $\theta_1$ with $\theta_2=0.5$ (left) and $\theta_2=3$ (right) and its evolution with depth for $\sigma_w^2 = \frac{1.6}{M}$.  The red line shows the fully-connected kernel (without averaging).}
   \label{fig:ill1}
\end{figure}

\paragraph{Illustrating the convolutional kernel}
Consider the kernel as defined in \eqref{eq:kernelgp}. As mentioned in e.g. \cite{lee17}, as the number of layers grows, the recurrent relation in \eqref{eq:fingp} for a fully-connected neural network approaches a fixed point so that the covariance function becomes a constant or piecewise constant map. We can illustrate this for a convolutional network by considering the kernel as a function of $\theta$, the angle between two vectors over which the dot-product in the convolution is computed. In the convolutional network the kernel is a sum, or average, over transforms of the kernels in the previous layer. In this way one would expect --due to the averaging structure-- a faster convergence to the flat structure where the structure itself depends on the the angles between the vectors over which the shifting dot-product is computed. For the sake of illustration consider $M=2$ and two distinct angles $\theta_1$ and $\theta_2$. We compute $K^2$ as a sum over the kernels in the previous layers $K^1(\theta_1)$ and $K^1(\theta_2)$, see Figure \ref{fig:ill1}, from which we conclude that the averaging indeed results in a more flat structure of the kernel. Compared to the fully-connected network in which the flattening depends on the depth, in convolutional networks it is additionally amplified by the convolutional structure, i.e. the averaging over the kernels.

\paragraph{Summary of the similarities and differences between a CNN and a FNN}
In a fully-connected network each hidden unit (for layer $l>1$) computes a summation of identically distributed variables, i.e. the sum of the weighted output from the previous layer, which are in a finite setting uncorrelated but not independently distributed. In a convolutional neural network each hidden unit computes a linear combination of not necessarily identically distributed (due to the covariance function being different for different terms in the summation) and uncorrelated but not independent variables. 


\section{Numerical results}\label{sec4}
In this section we numerically study and evaluate the conditions under which the behavior of the CNN output tends to that of a GP. In particular we study the discrepancy between the finite deep CNN and the corresponding Gaussian process in which we use the recurrent relationship for the kernel given in \eqref{eq:kernelgp}, through several intuitive examples. While our focus here is mostly on time series data and forecasting, the overall conclusions generalize to all prediction tasks.

The empirical discrepancy is determined using the maximum mean discrepancy (MMD) introduced in \cite{gretton12} and applied to measuring the similarity between GPs and fully-connected neural networks in \cite{matthews18}. The MMD between two distributions $\mathcal{P}$ and $\mathcal{Q}$ is defined as
\begin{align}
\mathcal{MMD}(\mathcal{P},\mathcal{Q},\mathcal{H}):=\sup_{||h||_{\mathcal{H}\leq 1}}\left[\mathbb{E}^\mathcal{P}[h]-\mathbb{E}^\mathcal{Q}[h]\right],
\end{align}
for which we use the unbiased estimator of squared MMD given in equation (3) in\\ \cite{gretton12}. For the empirical computations we take an input $X$, pass this through 500 convolutional neural networks and draw 500 samples from the GP with the covariance kernel corresponding to the depth of the CNN as defined in \eqref{eq:kernelgp}, and compare the results for different network parameters using the unbiased MMD estimator. We plot for comparison also the MMD estimator between two Gaussian processes with RBF kernels with length scales $l=\sqrt{2}$ and $l=4\sqrt{2}$.

\begin{figure}[H]
\centering
\includegraphics[scale=0.4]{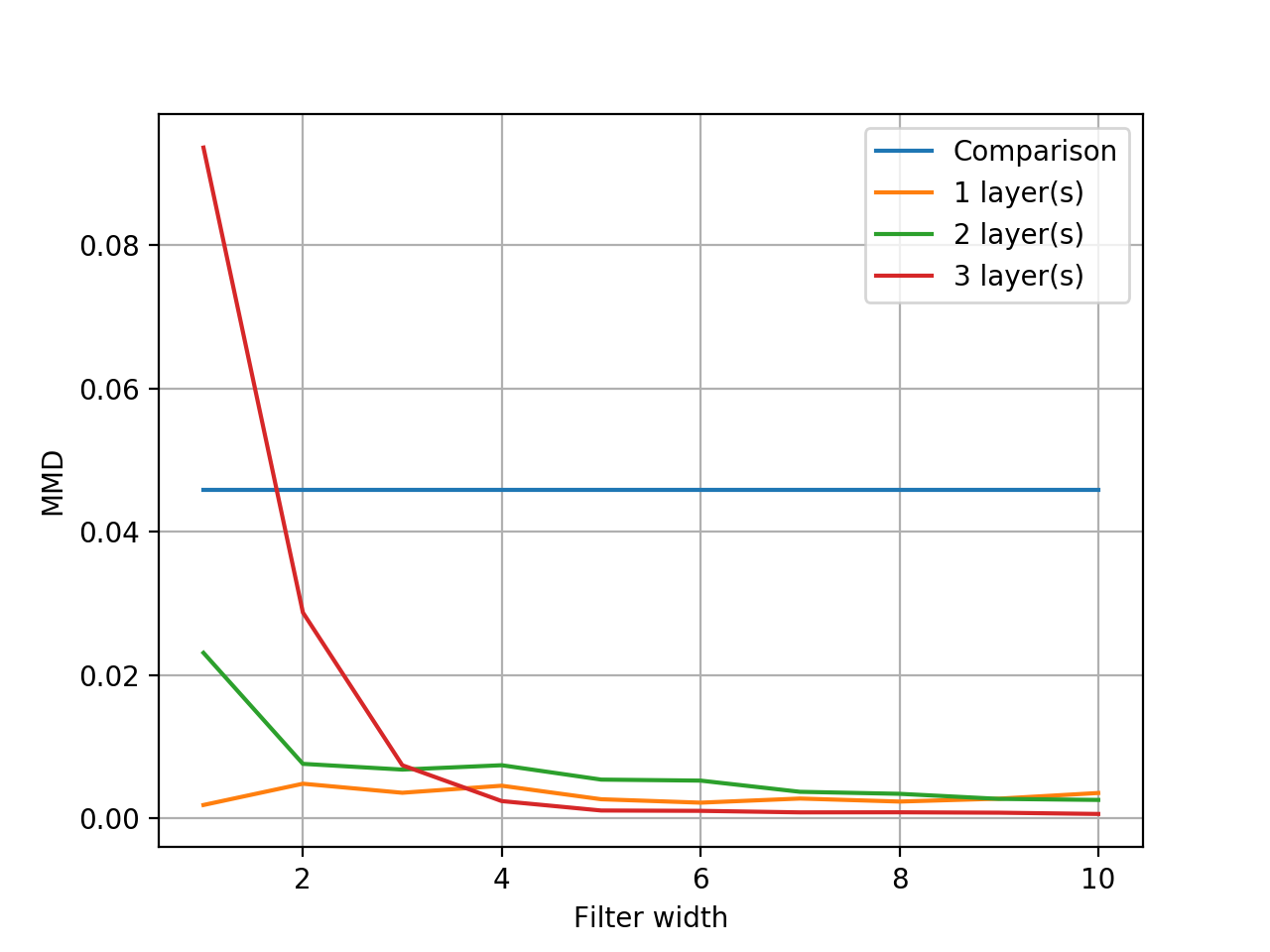}
\includegraphics[scale=0.4]{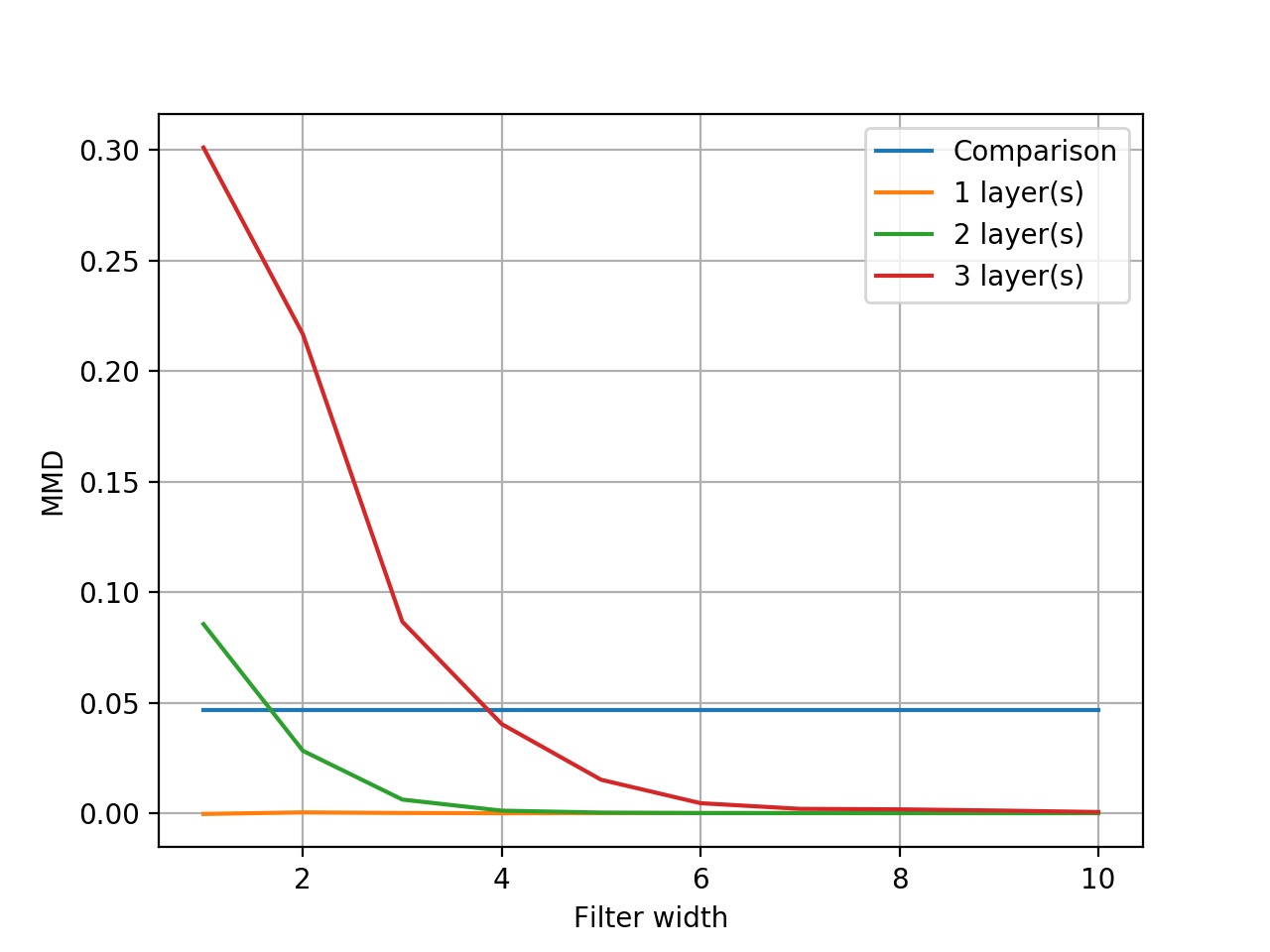}
\includegraphics[scale=0.4]{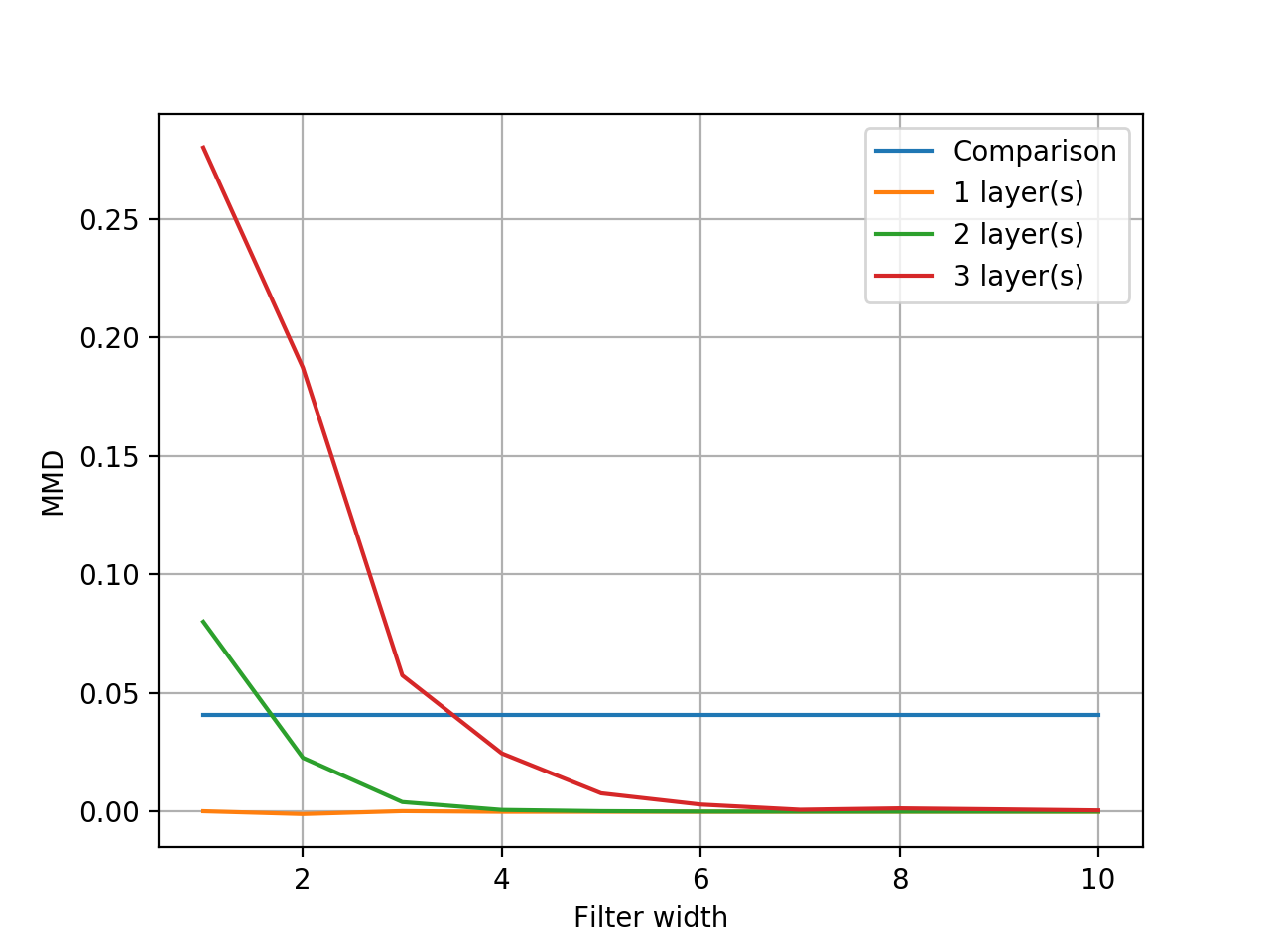}
\includegraphics[scale=0.4]{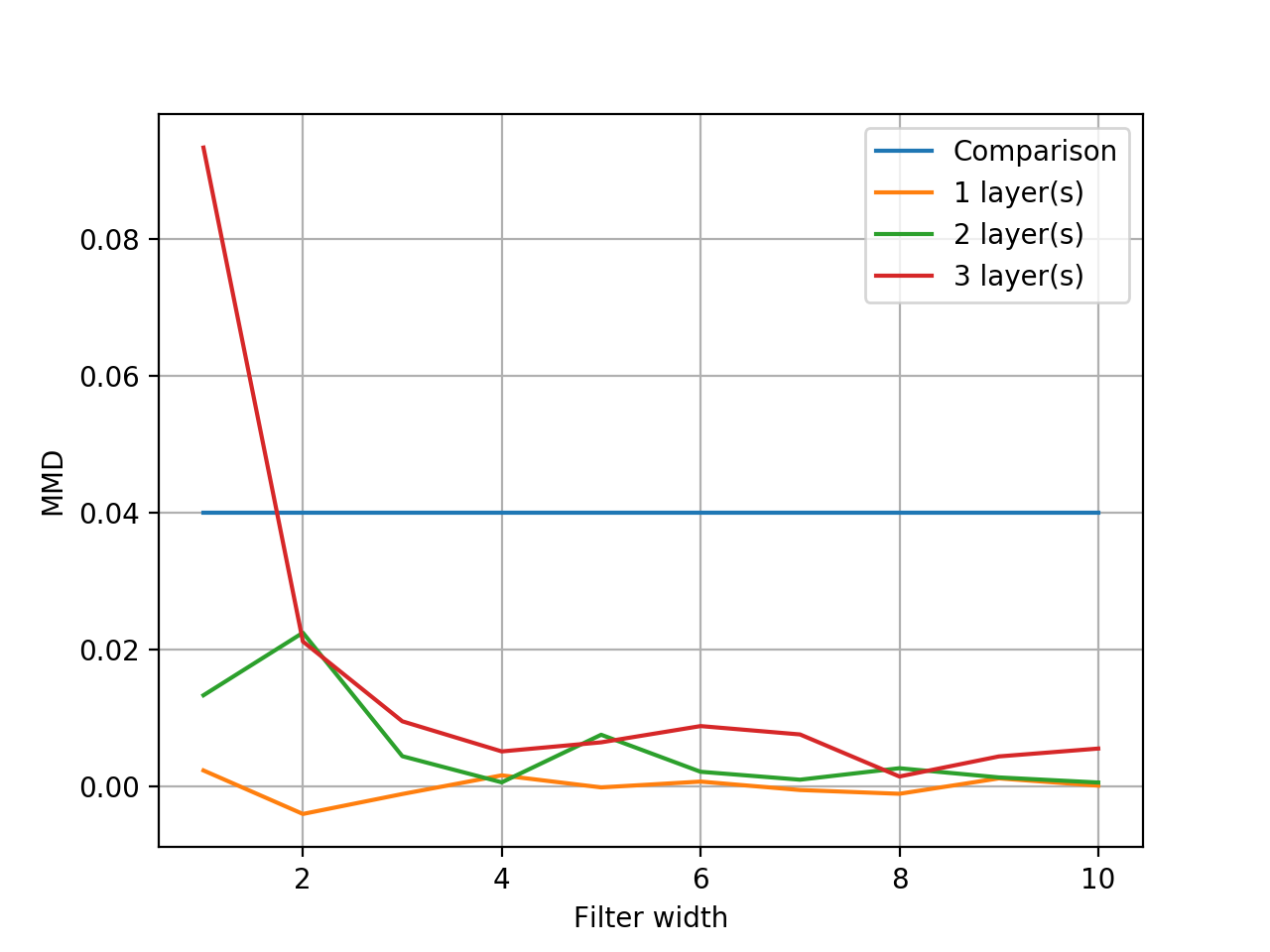}
\caption{MMD between the GP and CNN with a linear activation (top left), with a ReLU activation using \eqref{eq:kernelgp} (top right) and a ReLU (bottom left) and hyperbolic tangent (bottom right) using the MC evaluation of \eqref{eq:fingp} for not identically distributed inputs.}\label{fig:fig3}
\end{figure}

\paragraph{Prior discrepancy}
Let $x_i\overset{iid}{\sim}\mathcal{N}(0,1)$ for $i=1,...,d$ with $d=50$ and let $\sigma_w^2=1$. When computing the covariance matrix we condition on the input so that we have $K^1_{i,k}=\sigma_w^2 \sum_{j=1}^M x_{i-j}x_{k-j}$ so that the terms in the convolutional sum are not necessarily identically distributed; we compute the covariance function using \eqref{eq:kernelgp} and using a Monte Carlo (MC) evaluation of the expected value in \eqref{eq:fingp} by using 1000 Monte Carlo simulations in which we sample the weights $w_i\overset{iid}{\sim}\mathcal{N}(0,\sigma_w^2)$. Figures \ref{fig:fig3}-\ref{fig:fig3a} shows the results of the discrepancy between the GP and CNN. In the linear case we observe a fast convergence, but the discrepancy increases as we increase the number of layers, as is in accordance with the fact that the variables are not fully independent. Using a ReLU activation we observe a slower convergence in $M$ as we increase $l$, but nevertheless GP behavior is reached already for small filter sizes. For the bounded, symmetric hyperbolic tangent activation function the discrepancy between the GP and CNN output priors seems to be low regardless of the number of layers. 

We furthermore note that the expression for the kernel in \eqref{eq:kernelgp} is computed under the assumption of a sufficiently normal output in each layer, which in the GP, similar to the fully-connected network, is only attained in the limit. Nevertheless, compared to the MC computation of the kernel, the discrepancy between the two in the output prior seems to be sufficiently small, so that the analytical expression in \eqref{eq:kernelgp} is also able to mimic the recurrent relation in the deep convolutional network. When using CNNs for sequence forecasting a common setup is using a filter of size 2-8 and 2-5 layers depending on the required receptive field, see e.g. \cite{bai18}, \cite{borovykh17}. From Figure \ref{fig:fig3} one can conclude that the CNN indeed tends to behave like a Gaussian process, nevertheless when using a ReLU activation this behavior can be more or less avoided for a large number of layers and a relatively small filter size.


\begin{figure}[H]
\centering
\includegraphics[scale=0.4]{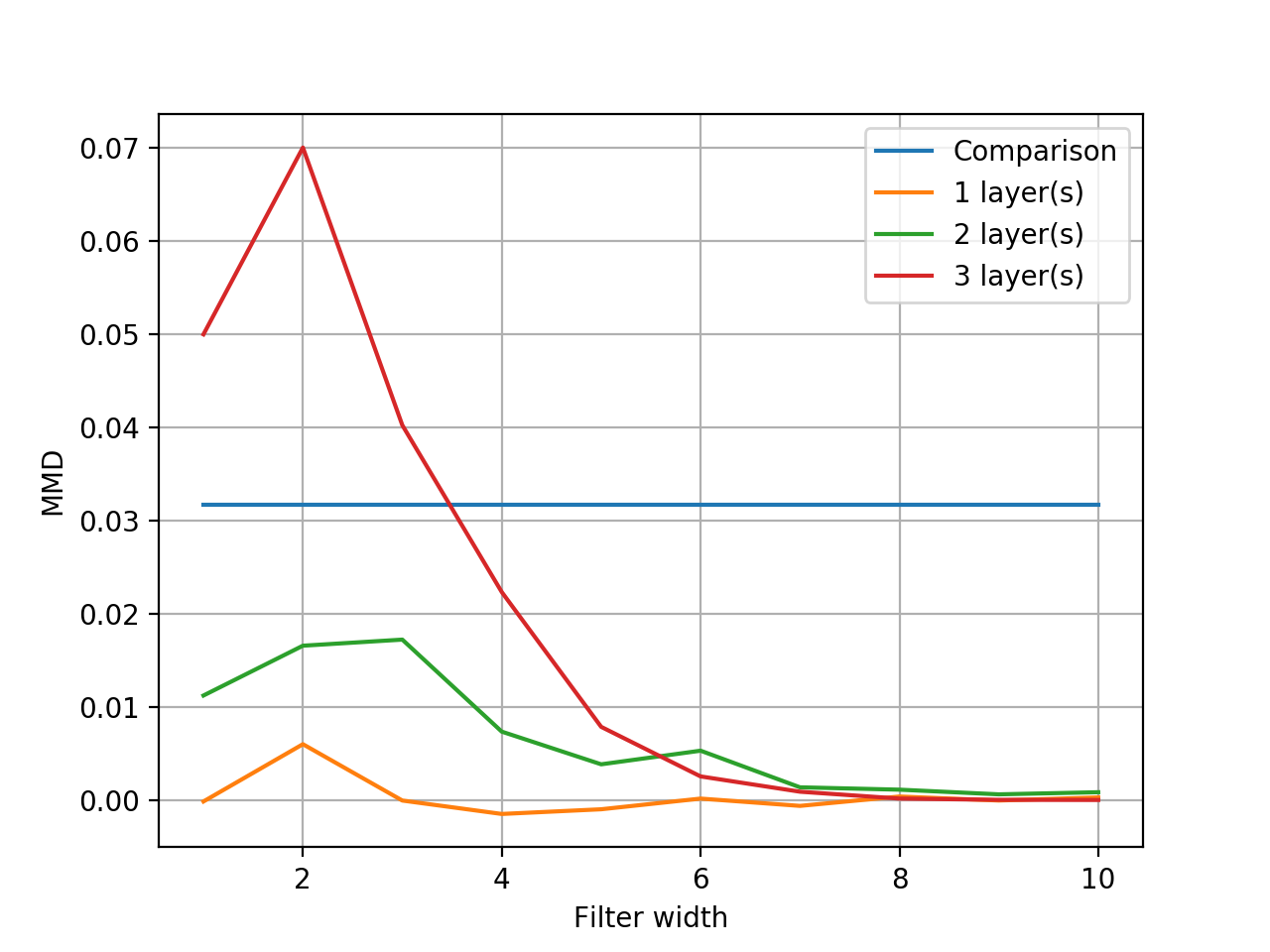}
\includegraphics[scale=0.4]{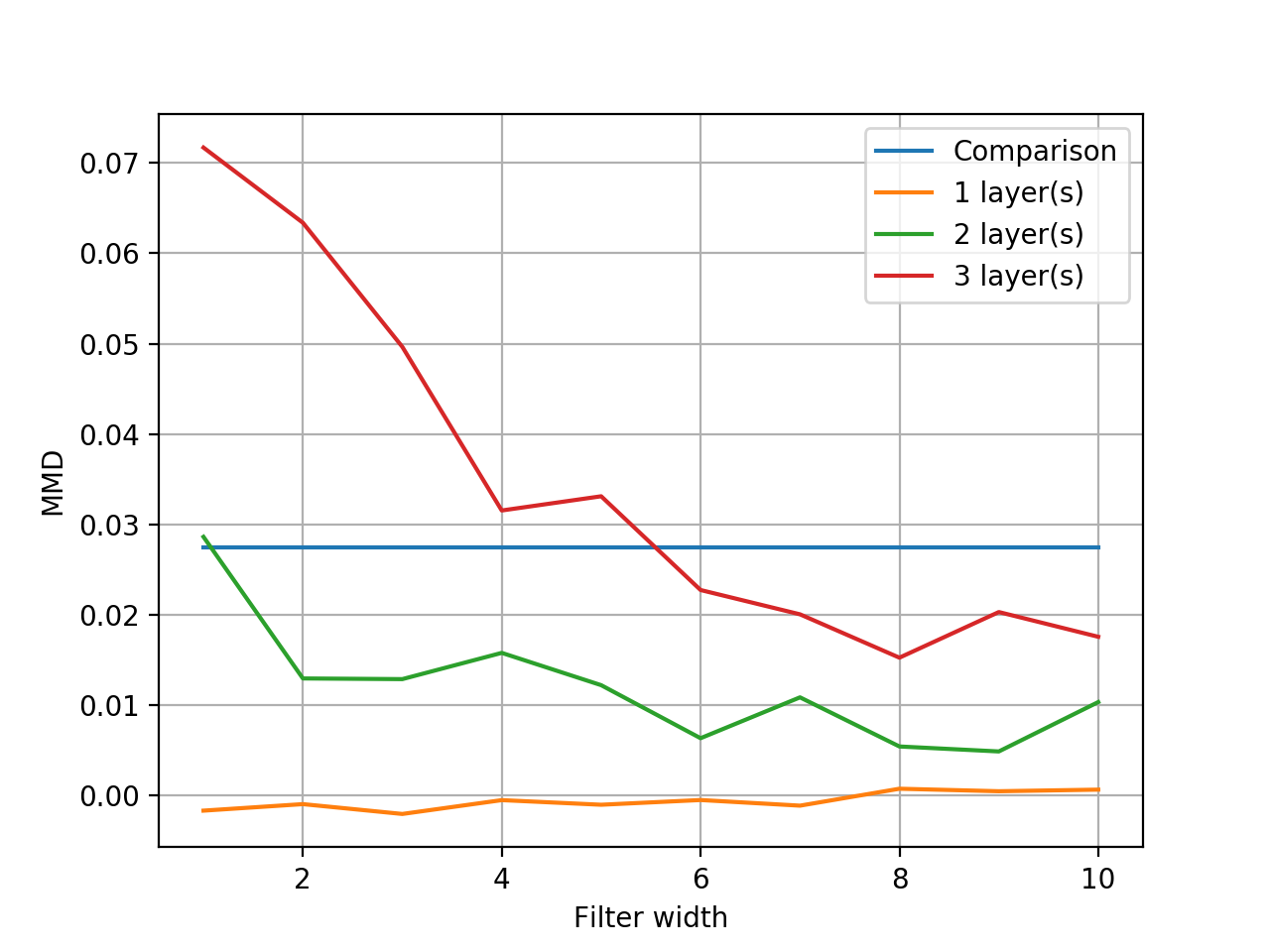}
\includegraphics[scale=0.4]{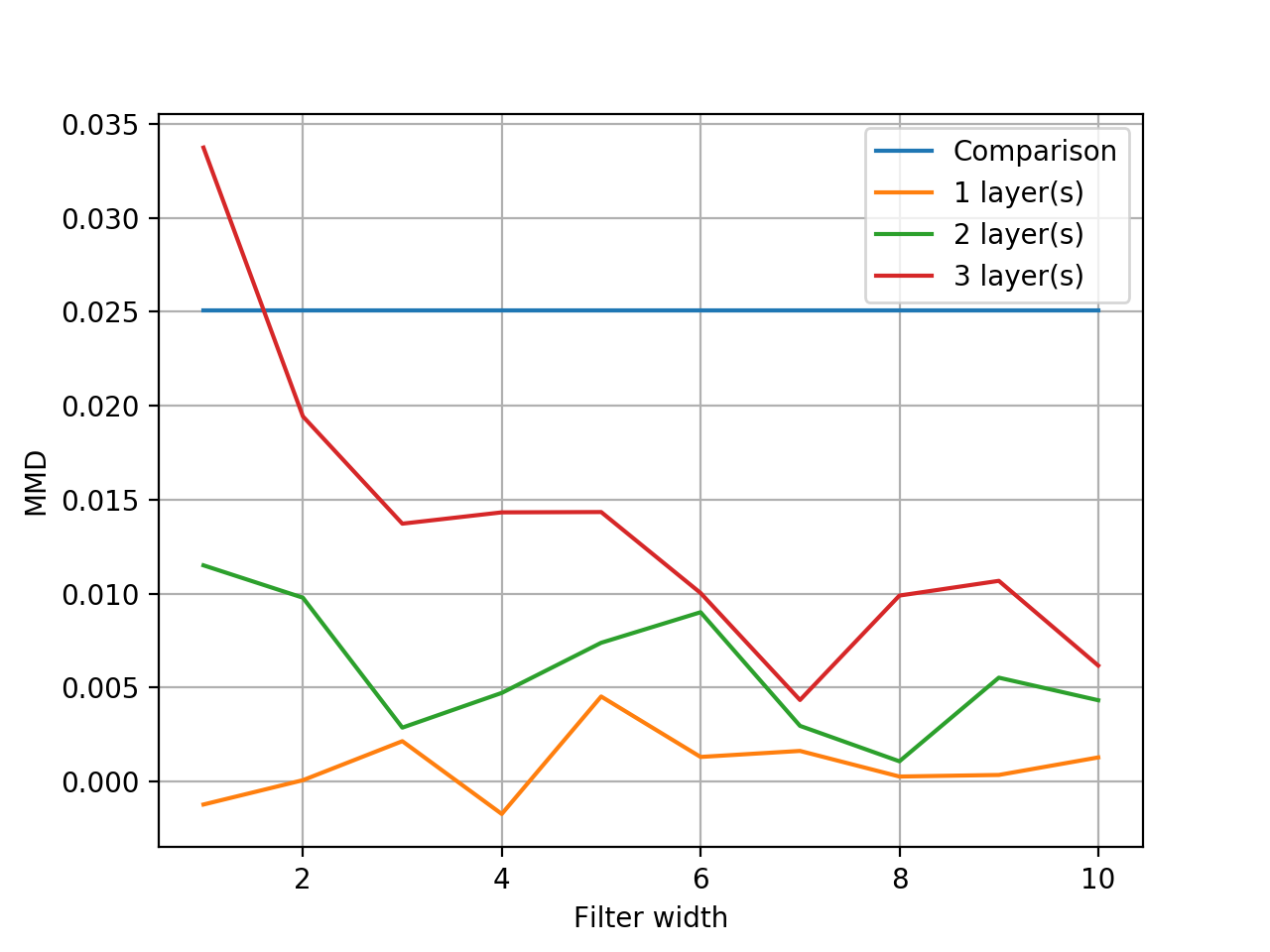}
\caption{MMD between the GP and CNN with linear (top left), with ReLU (top right), hyperbolic tangents (bottom) for an AR(2) series $x_i = \phi_1x_{i-1}+\phi_2x_{i-2}$ with coefficients $\phi_1=-0.6$ and $\phi_2=0.2$ and $d=50$.}\label{fig:fig4a}
\end{figure}

In Figure \ref{fig:fig3} we considered as input i.i.d. random variables, for which we showed that with a rectified linear unit the discrepancy between the CNN and GP priors increases as more layers are added, while for the hyperbolic tangent the difference remains small. Consider now $x_i$ to be a dependent autoregressive function, as is more realistic when working with time series, with finite covariance and let as before $\sigma_w^2=1$ and $d=50$. In Figure \ref{fig:fig4a} we show discrepancy between the CNN and GP output priors for the linear and non-linear activation functions. We note that for both the linear and ReLU activations, both of which are \emph{unbounded}, the CNN prior discrepancy with the GP prior is significantly higher than for the \emph{bounded} hyperbolic tangent activation. As expected the divergence increases with the number of layers due to increased dependency in the random variables.

\begin{figure}[h]
\centering
\includegraphics[scale=0.4]{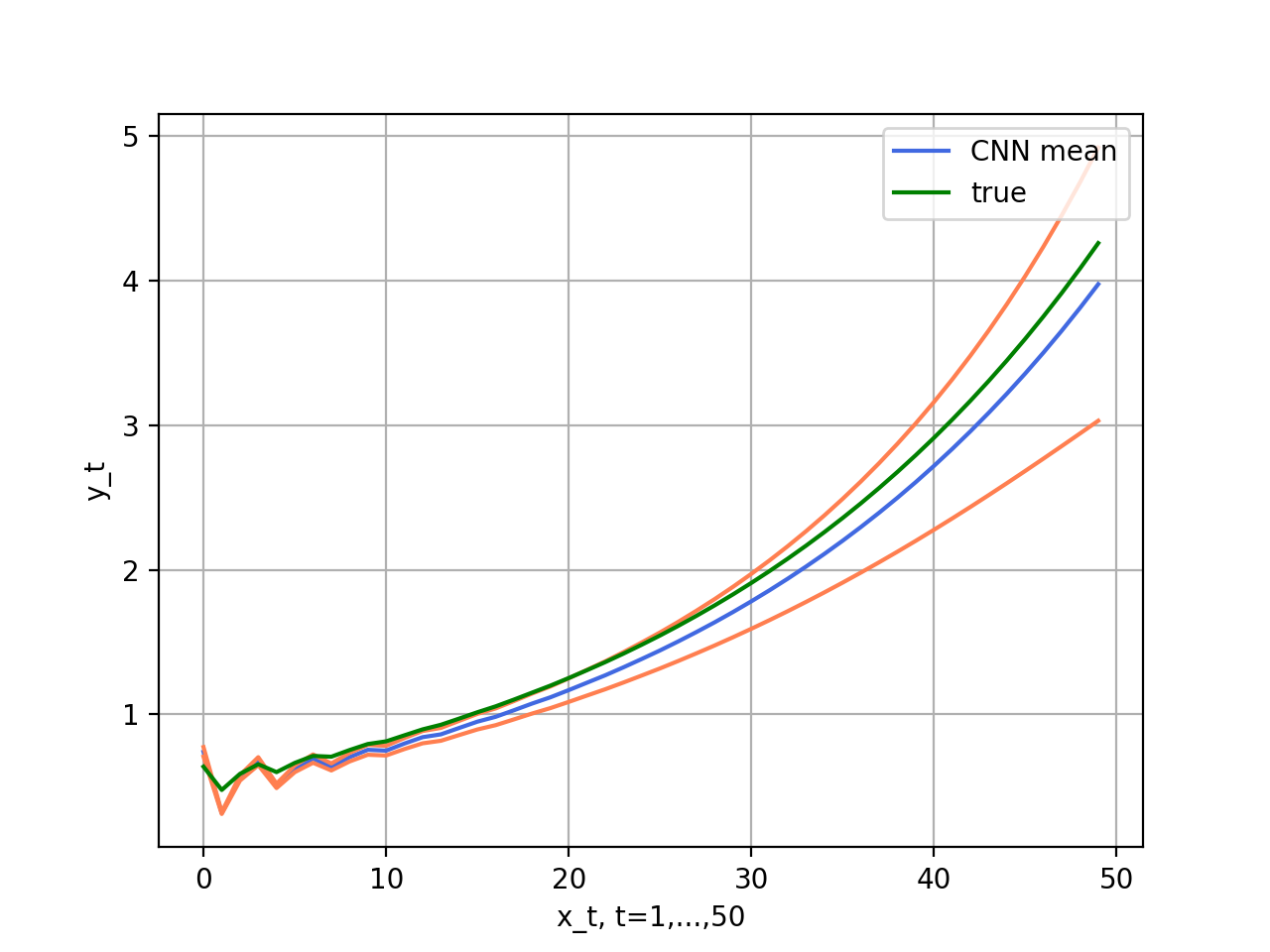}
\includegraphics[scale=0.4]{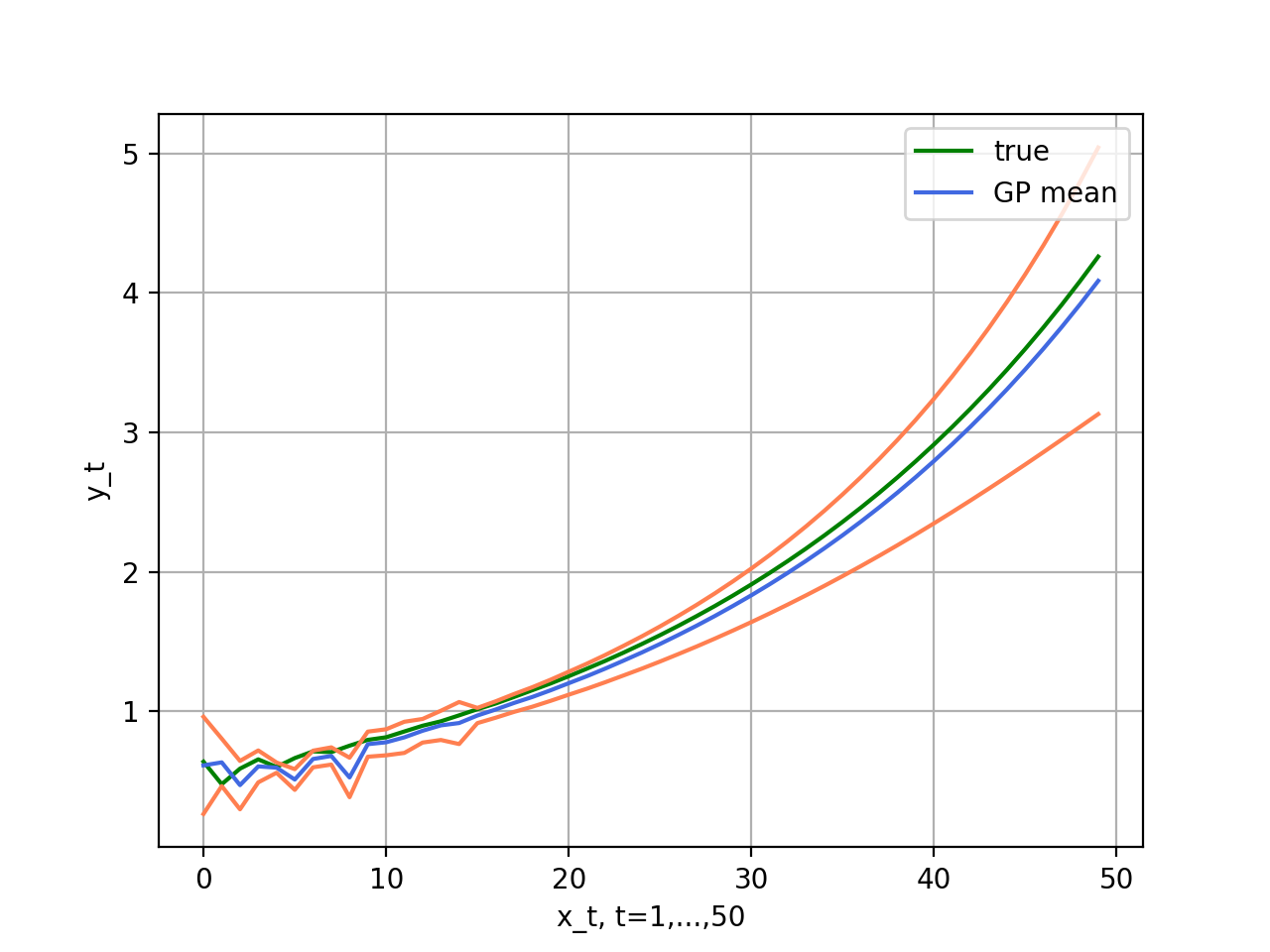}
\caption{A comparison between the trained CNN mean (left) and posterior inference in the corresponding Gaussian process (right) with credible confidence intervals. The CNN has one hidden layer and a filterwidth of three. }\label{fig:fig4}
\end{figure}

\paragraph{Posterior discrepancy}
Consider now $x_i$ to be an autoregressive time series of order three for $i=1,...,d$, a time series on which the CNN is well-able to learn the corresponding dynamics, let $\sigma_w^2 = 1$ and compute the covariance conditional on the input. Figure \ref{fig:fig4} shows the posterior GP mean and variance as well as the mean and variance of the CNN output obtained by training $N=100$ CNNs with random initialization (e.g. a simplified form of uncertainty estimation with an ensemble \cite{blundell16}; the network is trained by minimizing the mean squared error which is not a completely valid metric for capturing the uncertainty, nevertheless in our simplified example is enough to at least show the discrepancy between the posteriors). As expected, we conclude that if the priors of a GP and CNN are corresponding also their posteriors will be close.

%
 
\section{Conclusion and discussion}
In this paper we studied the convolutional neural networks by considering the deep network from a Gaussian process perspective. We extended the state of knowledge about GPs and deep networks to the convolutional network. In particular, using a Lyapunov-type bound for the case in which the inputs are not identically distributed we are able to obtain a CLT-like result on the discrepancy between the CNN and GP for one layer. For further layers, due to the dependency introduced by the layer outputs which are multivariate normally distributed the general form of the CLT for independent variables no longer applies. The empirical tests however show that using both linear and non-linear activations results in situations in which the CNN tends to behave like a GP. From a Bayesian perspective, if the priors of the two are indeed close enough we can use the analytical properties of GPs to obtain the uncertainty estimates in the CNN, thereby avoiding computationally intensive options such as Markov Chain Monte Carlo methods. 

Nevertheless, an important question remaining for further research is proving this convergence rate rigorously for the dependent variables in the summation in each layer.

Concluding, casting the network into a known framework such as the GPs provides interesting insights into the workings of the CNN, such as its distribution and the properties that can allow its output distribution to be close to a Gaussian -- in order to e.g. analytically compute the posterior distribution and use the GP uncertainty quantification--  and conversely the properties that allow to deviate from a Gaussian and in turn learn other, more complex distributions. This work was to the best of our knowledge the first attempt at doing so for the convolutional network structure, which possesses a very different structure compared to the fully-connected neural network. 


\bibliographystyle{apalike}
\bibliography{biblio}

\begin{thebibliography}{}

\bibitem[Bai et~al., 2018]{bai18}
Bai, S., Kolter, J.~Z., and Koltun, V. (2018).
\newblock Convolutional sequence modeling revisited.
\newblock {\em ICLR Workshop Track}.

\bibitem[Bentkus, 2005]{bentkus05}
Bentkus, V. (2005).
\newblock A {L}yapunov-type bound in $\mathbb{R}^d$.
\newblock {\em Theory of Probability \& Its Applications}, 49(2):311--323.

\bibitem[Borovykh et~al., 2017]{borovykh17}
Borovykh, A., Bohte, S., and Oosterlee, C.~W. (2017).
\newblock Conditional time series forecasting with convolutional neural
  networks.
\newblock {\em arXiv preprint arXiv:1703.04691}.

\bibitem[Bradshaw et~al., 2017]{bradshaw17}
Bradshaw, J., Matthews, A. G. d.~G., and Ghahramani, Z. (2017).
\newblock Adverial examples, uncertainty and transfer testing robustness in
  gaussian process hybrid deep networks.
\newblock {\em ArXiv}.

\bibitem[Calandra et~al., 2016]{calandra16}
Calandra, R., Peters, J., Rasmussen, C.~E., and Deisenroth, M.~P. (2016).
\newblock Manifold gaussian processes for regression.
\newblock In {\em Neural Networks (IJCNN), 2016 International Joint Conference
  on}, pages 3338--3345. IEEE.

\bibitem[Cho and Saul, 2009]{saul09}
Cho, Y. and Saul, L.~K. (2009).
\newblock Kernel methods for deep learning.
\newblock In {\em Advances in neural information processing systems}, pages
  342--350.

\bibitem[Damianou and Lawrence, 2013]{damianou13}
Damianou, A.~C. and Lawrence, N.~D. (2013).
\newblock Deep gaussian processes.
\newblock {\em International Conference on Artificial Intelligence and
  Statistics (AISTATS)}.

\bibitem[Durrande et~al., 2012]{durrande12}
Durrande, N., Ginsbourger, D., and Roustant, O. (2012).
\newblock Additive covariance kernels for high-dimensional {G}aussian process
  modeling.
\newblock {\em Annales de la Facult\'e de Sciences de Toulouse}, 21:481.

\bibitem[Duvenaud et~al., 2011]{duvenaud11}
Duvenaud, D.~K., Nickisch, H., and Rasmussen, C.~E. (2011).
\newblock Additive {G}aussian processes.
\newblock In {\em Advances in neural information processing systems}, pages
  226--234.

\bibitem[Gibbs, 1998]{gibbs98}
Gibbs, M.~N. (1998).
\newblock {\em Bayesian Gaussian processes for regression and classification}.
\newblock PhD thesis, University of Cambridge Cambridge, England.

\bibitem[Gretton et~al., 2012]{gretton12}
Gretton, A., Borgwardt, K.~M., Rasch, M.~J., Sch{\"o}lkopf, B., and Smola, A.
  (2012).
\newblock A kernel two-sample test.
\newblock {\em Journal of Machine Learning Research}, 13(Mar):723--773.

\bibitem[Hazan and Jaakkola, 2015]{hazan15}
Hazan, T. and Jaakkola, T. (2015).
\newblock Steps toward deep kernel methods from infinite neural networks.
\newblock {\em arXiv preprint arXiv:1508.05133}.

\bibitem[Hinton and Salakhutdinov, 2008]{hinton08}
Hinton, G.~E. and Salakhutdinov, R.~R. (2008).
\newblock Using deep belief nets to learn covariance kernels for gaussian
  processes.
\newblock In {\em Advances in neural information processing systems}, pages
  1249--1256.

\bibitem[Karpathy et~al., 2014]{karpathy14}
Karpathy, A., Toderici, G., Shetty, S., Leung, T., Sukthankar, R., and Fei-Fei,
  L. (2014).
\newblock Large-scale video classification with convolutional neural networks.
\newblock In {\em Proceedings of the IEEE conference on Computer Vision and
  Pattern Recognition}, pages 1725--1732.

\bibitem[Kim, 2014]{kim14}
Kim, Y. (2014).
\newblock Convolutional neural networks for sentence classification.
\newblock {\em arXiv preprint arXiv:1408.5882}.

\bibitem[Krizhevsky et~al., 2012]{krizhevsky12}
Krizhevsky, A., Sutskever, I., and Hinton, G.~E. (2012).
\newblock {ImageNet Classification with Deep Convolutional Neural Networks}.
\newblock {\em Advances in Neural Information Processing Systems 25}, pages
  1097--1105.

\bibitem[Lakshminarayanan et~al., 2016]{blundell16}
Lakshminarayanan, B., Pritzel, A., and Blundell, C. (2016).
\newblock Simple and scalable predictive uncertainty estimation using deep
  ensembles.
\newblock {\em arXiv preprint arXiv:1612.01474}.

\bibitem[Lee et~al., 2018]{lee17}
Lee, J., Bahri, Y., Novak, R., Schoenholz, S.~S., Pennington, J., and
  Sohl-Dickstein, J. (2018).
\newblock Deep neural networks as gaussian processes.
\newblock {\em Submitted to ICML}.

\bibitem[Mandt et~al., 2017]{mandt17}
Mandt, S., Hoffman, M.~D., and Blei, D.~M. (2017).
\newblock Stochastic gradient descent as approximate bayesian inference.
\newblock {\em arXiv preprint arXiv:1704.04289}.

\bibitem[Matthews et~al., 2018]{matthews18}
Matthews, A.~G., Hron, J., Rowland, M., Turner, R.~E., and Ghahramani, Z.
  (2018).
\newblock Gaussian process behaviour in wide deep neural networks.
\newblock {\em ICML}.

\bibitem[Neal, 2012]{neal12}
Neal, R.~M. (2012).
\newblock {\em Bayesian learning for neural networks}, volume 118.
\newblock Springer Science \& Business Media.

\bibitem[Rasmussen and Williams, 2006]{rasmussen06}
Rasmussen, C.~E. and Williams, C.~K. (2006).
\newblock {\em Gaussian processes for machine learning}, volume~1.
\newblock MIT press Cambridge.

\bibitem[{van den Oord} et~al., 2016]{vanoord16}
{van den Oord}, A., {Dieleman}, S., {Zen}, H., {Simonyan}, K., {Vinyals}, O.,
  {Graves}, A., {Kalchbrenner}, N., {Senior}, A., and {Kavukcuoglu}, K. (2016).
\newblock {WaveNet: A Generative Model for Raw Audio}.
\newblock {\em ArXiv e-prints}.

\bibitem[van~der Wilk et~al., 2017]{rasmussen17}
van~der Wilk, M., Rasmussen, C.~E., and Hensman, J. (2017).
\newblock Convolutional gaussian processes.
\newblock In {\em Advances in Neural Information Processing Systems}, pages
  2845--2854.

\bibitem[Williams, 1997]{williams97}
Williams, C.~K. (1997).
\newblock Computing with infinite networks.
\newblock In {\em Advances in neural information processing systems}, pages
  295--301.

\bibitem[Wilson and Adams, 2013]{wilson13}
Wilson, A. and Adams, R. (2013).
\newblock Gaussian process kernels for pattern discovery and extrapolation.
\newblock In {\em International Conference on Machine Learning}, pages
  1067--1075.

\bibitem[Wilson et~al., 2016]{wilson16}
Wilson, A.~G., Hu, Z., Salakhutdinov, R.~R., and Xing, E.~P. (2016).
\newblock Stochastic variational deep kernel learning.
\newblock In {\em Advances in Neural Information Processing Systems}, pages
  2586--2594.

\bibitem[Wilson et~al., 2011]{wilson11}
Wilson, A.~G., Knowles, D.~A., and Ghahramani, Z. (2011).
\newblock Gaussian process regression networks.
\newblock {\em arXiv preprint arXiv:1110.4411}.

\end{thebibliography}

 \end{document}